\pdfoutput=1
\documentclass[5pt]{article}
  
\usepackage[letterpaper]{geometry}
\usepackage{spconf,amsmath,epsfig}
\usepackage{enumitem}

\usepackage{multirow}
\usepackage{verbatim}
\usepackage{subcaption} 
\usepackage{etoolbox}
\usepackage{color}

\usepackage[toc,page]{appendix}

\usepackage{fancyhdr}
\begin{document}\sloppy

\def\x{{\mathbf x}}
\def\L{{\cal L}}

\patchcmd{\thebibliography}
{\settowidth}
{\setlength{\parsep}{0.0pt}\setlength{\itemsep}{0pt plus 0.0pt}\settowidth}
{}{}

\title{Visual Attribute-augmented Three-dimensional  Convolutional Neural Network for Enhanced Human Action Recognition}
%
\name{Yunfeng Wang$^{\star}$, Wengang Zhou$^{\star}$, Qilin Zhang$^{\dagger}$, Houqiang Li$^{\star}$\thanks{This work is the extended version of \cite{wang2018enhanced}. This work was supported in part to Dr. Houqiang Li by 973 Program under contract No. 2015CB351803 and NSFC under contract No. 61390514, and in part to Dr. Wengang Zhou by NSFC under contract No. 61472378 and No. 61632019, the Fundamental Research Funds for the Central Universities, and Young Elite Scientists Sponsorship Program By CAST (2016QNRC001).}}
\address{$^{\star}$University of Science and Technology of China, Hefei, Anhui, China\\
$^{\dagger}$HERE Technologies, Chicago, Illinois, USA
}
%
%
%

\maketitle

\begin{abstract}
  Visual attributes in individual video frames, such as the presence of characteristic objects and scenes, offer substantial information for action recognition in videos. With individual 2D video frame as input, visual attributes extraction could be achieved effectively and efficiently with more sophisticated convolutional neural network than current 3D CNNs with spatio-temporal filters, thanks to fewer parameters in 2D CNNs. In this paper, the integration of visual attributes (including detection, encoding and classification) into multi-stream 3D CNN is proposed for action recognition in trimmed videos, with the proposed visual Attribute-augmented 3D CNN (A3D) framework. The visual attribute pipeline includes an object detection network, an attributes encoding network and a classification network. Our proposed A3D framework achieves state-of-the-art performance on both the HMDB51 and the UCF101 datasets. 
  \end{abstract}
  \section{Introduction}
  \label{sec:intro}
  Action recognition has been extensively studied in past few years~\cite{wang2011action,wang2013action,simonyan2014two,Tran_2015_ICCV,cai2016effective,carreira2017quo,wang2018weighted,duan2018joint,lv2018video,huang2018video,zang2018attention}. Among these methods, recognizing human actions in videos with convolutional neural networks (CNNs) has been a popular research topic~\cite{simonyan2014two,feichtenhofer16,wang2016temporal,carreira2017quo}, thanks to the recent successes of CNNs in various computer vision tasks \cite{ran2017hyperspectral,ran2017convolutional}. Typically, these methods incorporate 3D CNNs to capture spatio-temporal information, optionally with a separate optical flow stream to account for low-level motion. Due to the increased parameter size in 3D filters in 3D CNNs, 3D CNNs are typically much shallower than their 2D counterparts \cite{Tran_2015_ICCV}. However, a new 3D CNN named ``Two-Stream Inflated 3D ConvNet" (I3D) was recently proposed in ~\cite{carreira2017quo}, with much deeper but still computationally feasible network design. 
  \begin{figure}[t!]
    \centering
    \begin{subfigure}[b]{0.22\textwidth}
      \includegraphics[width=\textwidth]{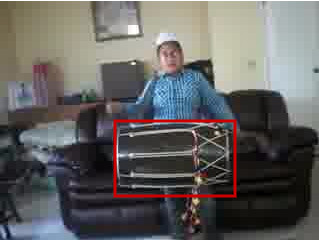}
      \caption{Playing Dhol}
    \end{subfigure}	\quad \begin{subfigure}[b]{0.22\textwidth}
      \includegraphics[width=\textwidth]{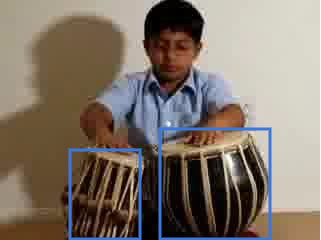}
      \caption{Playing Tabla}
    \end{subfigure}\\
  \caption{Sample video frames of (a) ``Playing Dhol" and (b) ``Playing Tabla" in UCF101 dataset, which confuse the I3D~\cite{carreira2017quo} algorithm. The visual attributes in bounding boxes (a detected dhol and $2$ tablas) could have helped to eliminate the ambiguity.}
  \label{fig:instruments_example}
  \end{figure}

 One of the potential improvements to current 3D CNN based action recognition systems could be the explicit inclusion of visual attributes as auxiliary information for recognition \cite{zhang2014can,zhang2015auxiliary,zhang2015multi}, which could be detected, encoded and classified with regular 2D CNNs effectively and efficiently. Visual attributes could play a vital role in some challenging action recognition cases, such as the one illustrated in Figure~\ref{fig:instruments_example}. The I3D~\cite{carreira2017quo} network struggles to distinguish ``Playing Dhol" from ``Playing Tabla". However, the visual attributes\footnote{Denoted in red and blue bounding boxes in Figure~\ref{fig:instruments_example}.} could have helped to eliminate such confusions. 
  \begin{figure*}[t]
    \centering
    \includegraphics[width=0.99\textwidth]{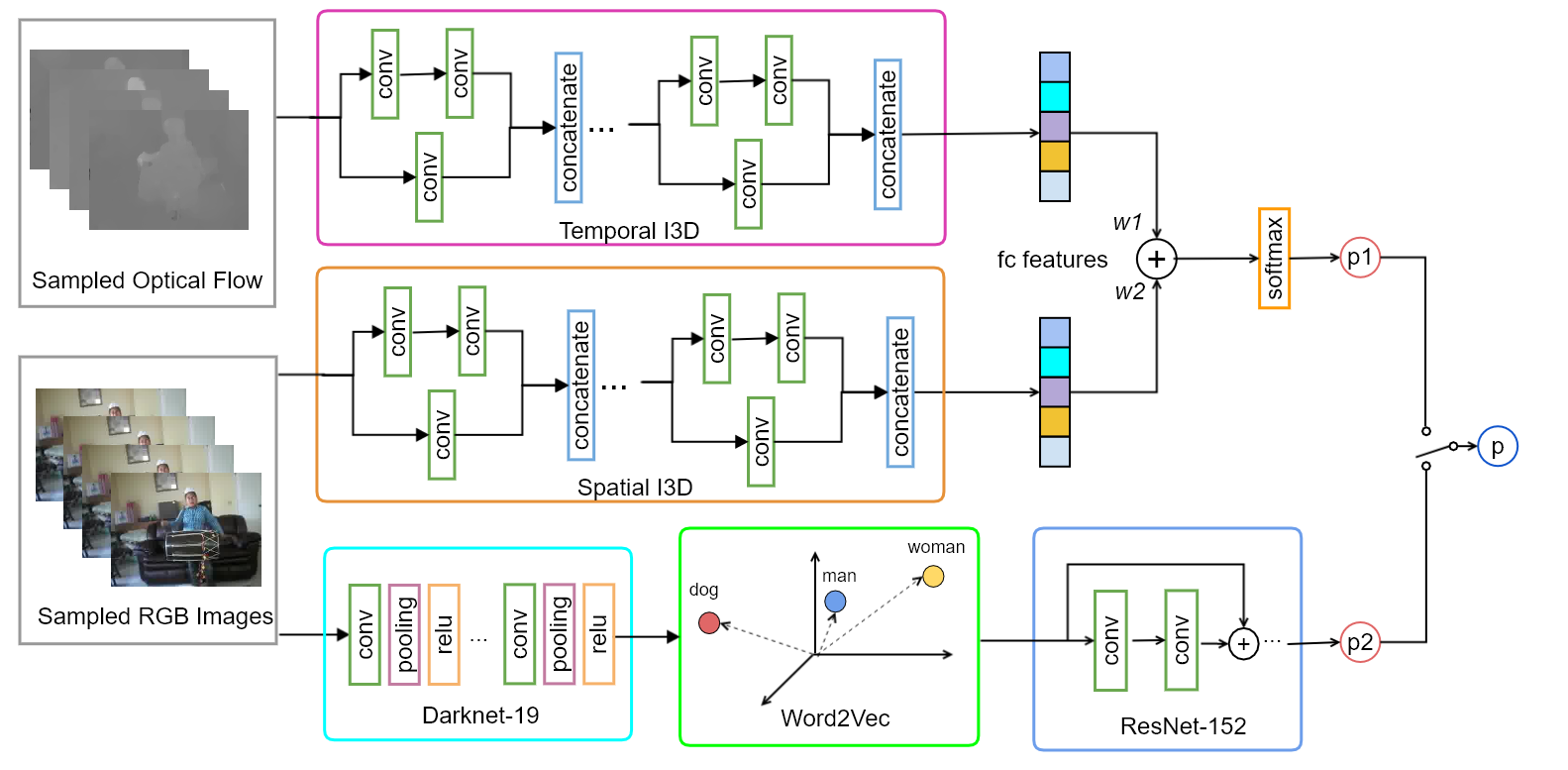}\\
    \caption{A3D framework overview. Optical flow and RGB inputs are fed to the temporal stream and spatial stream of the I3D ConvNet, respectively, followed by an early fusion stage which merges fc features and a softmax to obtain the 3D CNN pipeline prediction $p_{1}$. The visual attribute pipeline is illustrated at the bottom, with sampled RGB inputs fed to an object/attribute candidate’s detector network (e.g., Darknet-19 from YOLO9000), a video attribute representation constructor (e.g., Word2vec) and finally a classification network (finetuned ResNet-152). The prediction $p_{2}$ from this attribute pipeline is ultimately combined with $p_{1}$ by $p=p_{1}\times I(p_{1} - T) + p_{2} \times I(T-p_{1})$, with $I(x)$ being an indicator function and $T$ being a global threshold. $I(x)=1$ if $x>0$, and $I(x)=0$ if $x\leq 0$.}
    \label{fig:framework}
    \end{figure*}

  Based on this intuition, an enhanced action recognition framework is proposed, namely the visual Attribute-augmented 3D CNN (A3D), comprising both a 3D CNN pipeline and a visual attributes pipeline. The 3D CNN pipeline is a modified I3D network with temporally subsampled RGB/optical flow inputs, trained with filtered attributes. While the visual attributes pipeline consists of a YOLO9000~\cite{redmon2016yolo9000} visual attribute candidate's detector, a ResNet~\cite{he2016deep}$/$Word2Vec~\cite{mikolov2013efficient}$+$Mean-pool$/$NetVLAD~\cite{arandjelovic2016netvlad} attributes encoders and different classifiers. Based on the output probabilities from both pipelines, the final prediction is constructed by global thresholding and weighted summation, as illustrated in Figure~\ref{fig:framework}. 
  In order to achieve fair comparison, we downloaded the network model pre-trained on dataset ``Kinectics" and reimplemented the finetuning steps on both UCF101 and HMDB51 ourselves, and named this reimplemented version as I3D*\footnote{The reimplmented I3D* slightly underperforms the official I3D on both the UCF101 and the HMDB51 datasets.}, which is later used for comparison in Table~\ref{tab:comp_methods} of Section~\ref{sec:Exp}.
  \section{Proposed A3D Framework}
  As in Figure~\ref{fig:framework}, two pipelines are present in the A3D framework. The 3D CNN pipeline is illustrated on the top, with a temporal I3D stream and a spatial I3D stream, processing sampled optical flow images and regular RGB images, respectively. The fc feature outputs from both streams are merged before the softmax scoring layer (which is different from the original I3D implementation in \cite{carreira2017quo}) to obtain pipeline prediction $p_1$. On contrary, the visual attribute pipeline is illustrated at the bottom of Figure~\ref{fig:framework}, with only regular RGB images as inputs. This attribute pipeline contains a generic object (i.e., visual attribute candidate) detector (Darknet-19 from YOLO9000), a video attribute representation constructor (mean pooling, NetVLAD~\cite{arandjelovic2016netvlad} or Word2vec~\cite{mikolov2013efficient}) and finally a CNN based classifier, which produces the pipeline prediction $p_2$. As an auxiliary information source\footnote{Additional modality of information is demonstrated to be helpful in applications such as \cite{abeida2013iterative,zhang2012fast,zhang2011fast}, but the quality of the new modality need to be examined. Here we intuitively use the probability threshold to control the modality quality.}, the attribute pipeline output $p_2$ is only activated when $p_1$ falls below a predefined global threshold. Details of major components of A3D are provided as follows. 
  %
  %
  %
  %
  \subsection{Revised Two-stream Fusion in 3D CNN Pipeline}
  Specifically, the inflated Inception-V1 module proposed in~\cite{carreira2017quo} is used in both streams of the 3D CNN pipeline.  After finishing network training, the outputs of the last fully connected layer from both streams are weighted (with $w_1=0.6$ and $w_2=0.4$, respectively. Values of the weights are empirically determined.), element-wise summed, and fed to a softmax scoring layer to obtain prediction $p_1$. Unlike the original fusion scheme in ~\cite{carreira2017quo}, where fusion is carried out after the softmax scoring layer, the revised fusion strategy in the 3D CNN pipeline of A3D leads to performance enhancements, as later detailed in Table~\ref{tab:early_fusion_comp} of Section~\ref{sec:Exp}. 
  \subsection{Visual Attribute Candidate Detection}
  To exploit relevant visual attributes to action recognition tasks, Darknet-19 (from YOLO9000) object detector is applied on randomly selected video frames. Thanks to the very large number of object categories (approximately $9000$), this object detector could serve as an off-the-shelf generic attribute mining tool. In order to remove obvious outliers with minimal useful information, attributes candidates with detected bounding boxes smaller than $20$-pixels are removed. For an attribute $a$ (there are approximately $9000$ distinctive choices of $a$) selected from video of class $t$ ($t$ denotes the label of the video), we assign $t$ as the label of $a$. Attribute candidates are cropped from original images and saved for subsequent steps.   
  \subsection{Video Attribute Representation and Classification}
  \label{sec:encoding_classification}
  %
  %
  %
  %
  %
  %
  The number of attributes detected from different video is variant, thus it is necessary to encode them in a format (i.e., video attribute representation) with consistent size. Three strategies of attribute representation construction are tested. We find the strategy \textit{Word2Vec+ResNet-152 finetune} achieves the best accuracy. Please refer to Appendix for detailed experiment settings and results.
  \subsection{Joint Inference of Two Pipelines}
  In the testing phase, both pipelines jointly perform classification inference. Given a test video, $p_{1}$ and $p_{2}$ are obtained by the 3D CNN pipeline and visual attribute pipeline, respectively. Final prediction $p$ is obtained by, 
  \begin{equation}
  p = p_{1}\times I(p_{1} - T) + p_{2}\times I(T - p_{1}), \label{eq:combination}
  \end{equation}
  where $T$ is a global threshold of prediction confidence, empirically set to $0.1$, and $I(\cdot)$ is an indicator function, 
  \begin{equation}
  I(x) = \begin{cases}
  1 & \text{if } x>0 \\
  0 & \text{if } x\leq0 
  \end{cases}.
  \end{equation}
  If $p_{1}$ is larger than the threshold $T$, the 3D CNN pipeline is deemed confident enough and it determines the final prediction $p$. Otherwise, the visual attribute pipeline takes over and $p$ falls back on $p_2$. 
  %
  %
  %
  %
  %
  \section{Experiments} \label{sec:Exp}
  The proposed A3D framework is evaluated on both the HMDB51 and UCF101 datasets. the standard evaluation protocol is used and the average accuracies over three splits are reported (unless otherwise specified). 
  
  %
  In the following experiments, the pretrained (on Kinetics dataset) inflated Inception-V1 models are finetuned on UCF101 and HMDB51 dataset, separately. The temporal I3D stream and spatial I3D stream are also separately trained on optical flow images and RGB images, respectively. For each video, 64 regular RGB frames and 64 optical flow frames are randomly selected as the input to finetune the 3D CNN pipeline. Regular stochastic gradient descent (SGD) and back propagation (BP) are used to optimize training loss, with initial learning rate of $0.001$, a learning rate decay of $0.8\times$ every $10$ epochs and a total of $50$ epochs.

  In the original I3D implementation~\cite{carreira2017quo}, fusion is carried out after the softmax scoring layer. On contrary, the revised fusion in the proposed 3D CNN pipeline happens immediately after obtaining the features from fully connected layers. FC layer features are weighted by $w_1 = 0.4$ and $w_2 = 0.6$ (weight values are determined empirically) and summed element-wise.  As shown in Table~\ref{tab:early_fusion_comp} (action classification based purely on $p_1$), the revised fusion outperforms original fusion on the ``split1" of both datasets. Therefore, revised fusion is used throughout the remaining parts of the Section~\ref{sec:Exp}. 
  \setlength{\tabcolsep}{6pt}
  \begin{table}[htbp]
    \centering
        \caption{3D CNN Pipeline: accuracies based purely on $p_1$, comparing $2$ fusion strategies.}
        \label{tab:early_fusion_comp}
        \begin{tabular}{|l|c|c|}
          \hline
          Method & UCF101(split1) &  HMDB51(split1) \\
          \hline
          Original Fusion  & 95.67\% & 79.00\% \\
          \hline
          Revised Fusion & \textbf{97.09\%} & \textbf{79.22\%} \\
          \hline
        \end{tabular}
  \end{table}

\begin{figure}[t]
  \centering
    \includegraphics[width=0.48\textwidth]{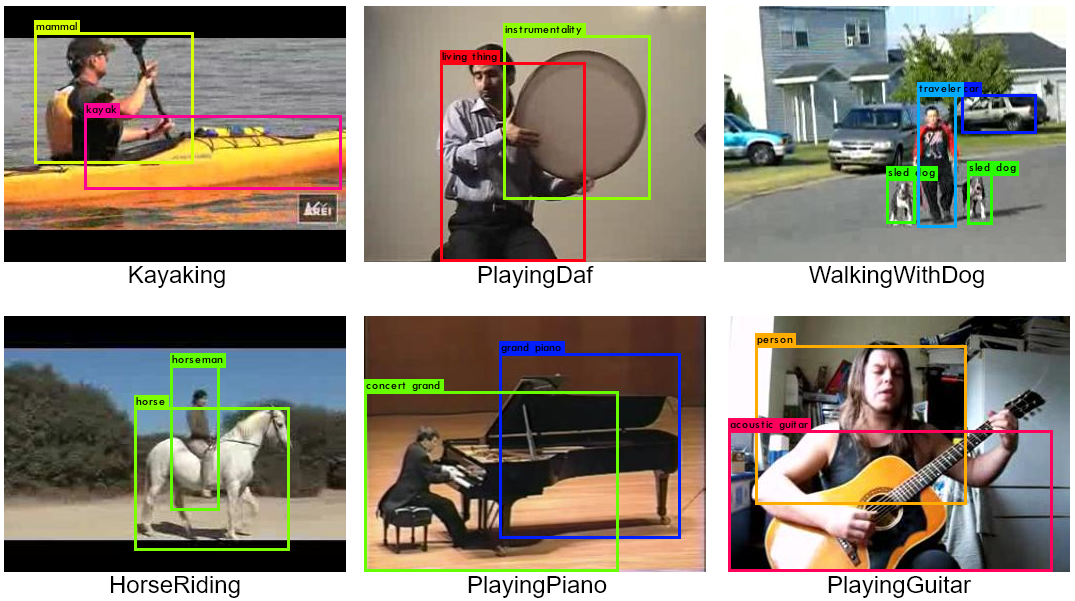}\\
    \caption{Sample results of visual attributes detection}  
    \label{fig:yolo9000_output}
  \end{figure}

  Since UCF101 and HMDB51 are trimmed video datasets, the context and scene information in each video is simple, thus we only randomly select one frame in each video for visual attribute extraction. We adapt the implementation of YOLO9000 in~\cite{redmon2016yolo9000} to a version compatible with parallel computing and set a relatively low threshold of $0.02$, to ensure plenty of video attribute candidates are detected in each video. Samples of video attribute candidates detection results on UCF101 are shown in Figure~\ref{fig:yolo9000_output}. 
  %
  %
  %
  %
  %
  %
  \setlength{\tabcolsep}{6pt}
  \begin{table}[t]
  \centering
        \caption{Comparison of 3D CNN pipeline $p_1$ based classifier, video attribute pipeline $p_2$ based classifier and $p$ based joint classifier on split1 of UCF101 and HMDB51.}
        \label{tab:comp_baseline}
        \begin{tabular}{|l|c|c|}
          \hline
          Classifier based on & UCF101(s1) &  HMDB51(s1) \\
          \hline
          3D CNN pipeline $p_1$ & 97.09\% & 79.22\% \\
          Attribute pipeline $p_2$ & 37.10\% & 27.05\% \\
          Joint A3D framework $p$ & \textbf{97.44\%} & \textbf{79.35\%} \\
          \hline
        \end{tabular}
  \end{table}
  \setlength{\tabcolsep}{7pt}
  \begin{table}[t]
    \centering
    \caption{Running time comparison on split1 of $2$ datasets}
    \label{tab:time_overhead}
    \begin{tabular}{|c|c|c|c|c|}
      \hline
      \multirow{2}{*}{Method} & \multicolumn{2}{c|}{UCF101(s1)} & \multicolumn{2}{c|}{HMDB51(s1)} \\\cline{2-3}\cline{4-5}
      & Training & Testing & Training & Testing \\ 
      \hline
      I3D    & 41.67h   & 0.53h  & 16.70h     & 0.22h \\
      A3D    & 45.32h    & 0.61h  & 17.82h   & 0.26h \\
      Overhead & 8.76\%  & 1.56\% & 6.71\%     & 1.82\%\\
      \hline
    \end{tabular}
  \end{table}
  %
  
  %
  \setlength{\tabcolsep}{10pt}
  \begin{table}[h!]
  \centering
        \caption{Comparison of A3D with competing methods}
        \label{tab:comp_methods}
        \begin{tabular}{|l|c|c|}
          \hline
          Method &  UCF101 &  HMDB51 \\
          \hline
          iDT~\cite{wang2013action} & 85.9\% & 57.2\% \\
          Two-Stream~\cite{simonyan2014two} & 88.0\% & 59.4\% \\
          C3D~\cite{Tran_2015_ICCV} & 85.2\% & - \\
          TDD + iDT~\cite{wang2015action} & 91.5\% & 65.9\% \\
          TSN~\cite{wang2016temporal} & 94.2\% & 69.4\% \\
          P3D ResNet + iDT~\cite{qiu2017learning} & 93.7\% & - \\
          ST-ResNet + iDT~\cite{feichtenhofer2017spatiotemporal} & 94.6\% & 70.3\% \\
          I3D* & 97.1\% & 79.4\% \\
          \hline
          Proposed A3D   & \textbf{97.4\%} & \textbf{80.5\%} \\
          \hline
        \end{tabular}
  \end{table}

  In order to investigate the individual contributions of the 3D CNN pipeline and visual attribute pipeline of the proposed A3D framework, an ablation study is carried out and the results are summarized in Table~\ref{tab:comp_baseline}. Although attribute pipeline $p_2$ based classifier substantially underperforms the counterpart based on $p_1$ (possibly due to excessive noises incurred by irrelevant objects), it stills helps to incorporate $p_2$ via a reasonable combination function (e.g., Eq.~\eqref{eq:combination}). The classifier based jointly on both pipelines $p$ achieves the highest accuracy on split1 of both datasets. 
  Computational complexity wise, the proposed A3D framework is only marginally heavier than the I3D algorithm. A comparison of running time\footnote{Based on our hardware setup, GPU: $1\times$ Nvidia GTX 1080Ti, CPU: $2\times$ Intel(R) Xeon(R) E5-2640 v4 2.40GHz and 512 GB memory.} (including both the training phase and testing phase with split1 of both datasets) is summarized in Table~\ref{tab:time_overhead}. The bottom row in Table~\ref{tab:time_overhead} contains the percentile overhead running time of A3D over I3D, which are all under $10\%$.  
  In Table~\ref{tab:comp_methods}, the end-to-end action recognition accuracies are compared over all $3$ splits of both datasets. The proposed A3D framework achieves the highest overall accuracies on both datasets. 
  %
  \section{Conclusions}
  In this paper, an enhanced action recognition framework is proposed, namely the visual Attribute-augmented 3D CNN (A3D), based on the explicit incorporation of visual attributes in addition to a two-stream 3D CNN pipeline. A new two-stream fusion strategy for the 3D CNN pipeline is introduced for improved performance. Different designs of multiple components (video attribute representation constructor, classifier) in the new video attribute pipeline are tested and empirically determined. An ablation study confirms the value of the additional video attribute pipeline. Overall, the proposed A3D framework outperforms all competing ones on two benchmarks. 

  \bibliographystyle{IEEEbib}
  \bibliography{A3D}

  \begin{appendices}
  \section{Video Attribute Representation and Classification}
\begin{figure}[h]
    \centering
      \begin{subfigure}[b]{0.47\textwidth}
        \includegraphics[width=\textwidth]{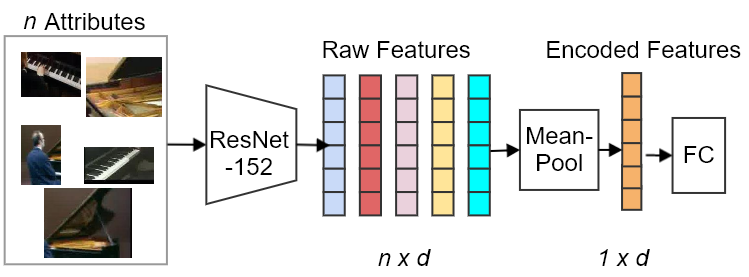}
        \caption{ImageNet pretrained ResNet-152 feature extractor, mean pooling for attribute representation construction, A fully connected layer for classification.}
      \end{subfigure}\\
      \begin{subfigure}[b]{0.47\textwidth}
        \includegraphics[width=\textwidth]{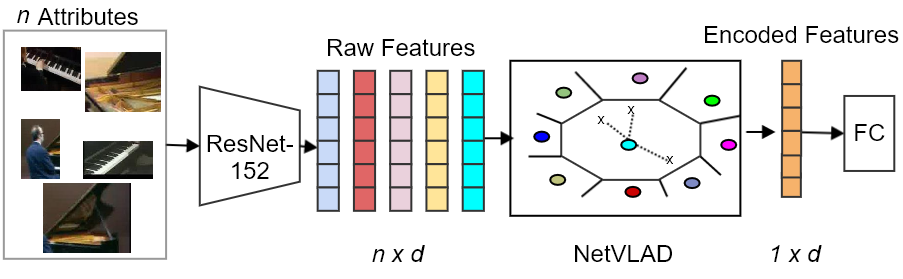}
        \caption{ImageNet pretrained ResNet-152 feature extractor, NetVLAD for attribute representation construction, A fully connected layer for classification.}
      \end{subfigure}\\
      \begin{subfigure}[b]{0.47\textwidth}
        \centering
        \includegraphics[width=0.7\textwidth]{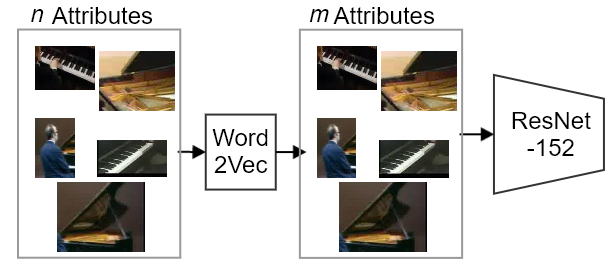}
        \caption{Word2vec for selection of the most informative attributes, a finetuned (with filtered attributes) ResNet-152 ConvNet for classification.}
      \end{subfigure}
      \caption{Video attribute representation and classification}
      \label{fig:encoding_classification}
    \end{figure}

  The number of attributes detected from different video is variant, thus it is necessary to encode them in a format (i.e., video attribute representation) with consistent size. As shown in Figure~\ref{fig:encoding_classification}, three strategies of attribute representation construction are tested.
    \begin{enumerate}[noitemsep,leftmargin=*,topsep=0pt]
      \item \textbf{ResNet-152 + Mean-Pool + FC}: Cropped visual attributes images are fed to a pretrained (on ImageNet) ResNet-152 network to extract features, followed by a mean pooling operation on all features extracted from the same particular video, to obtain a fixed-size video attribute representation. Attribute representations are subsequently classified by a fully connected layer. 
      \item \textbf{ResNet-152 + NetVLAD + FC}: Instead of mean pooling, a NetVLAD~\cite{arandjelovic2016netvlad} is used to aggregate all features from the same particular video. Other steps are identical to the previous \textit{ResNet-152 + Mean-Pool + FC} strategy. 
      \item \textbf{Word2Vec + ResNet-152 finetune}: Since the labels of YOLO9000 are based on WordNet~\cite{miller1995wordnet}, it is possible to reuse the labeling information of attributes and remove irrelevant attributes using natural language processing techniques. Firstly, attributes with the labels relevant to ``person" are removed, since all types of human action videos contain people thus no additional information is provided. Secondly, a pre-trained (on WordNet corpus) Word2vec model is used for removing rarely relevant attributes during the training phase. Each ground truth video label $t$ (containing a verb, e.g., ``play" and a noun, e.g., ``guitar") is disassembled into 2 words and fed to such Word2vec model to obtain a vector of score $\mathbf{s}(t)$. All extracted video attributes ($a_1$, $\cdots$, $a_N$) from the same video are separately fed to the same Word2vec model (to obtain $\mathbf{s}(a_1)$, $\cdots$, $\mathbf{s}(a_N)$). By comparing the cosine similarities between $\mathbf{s}(t)$ and $\mathbf{s}(a_n)$ with a predefined threshold\footnote{$T_{sim}$ is empirically set to $0.5$ in this paper.} $T_{sim}$, 
      \begin{equation}
      \text{SIM}\left(\mathbf{s}(t), \mathbf{s}(a_n)\right) = \frac{\mathbf{s}(t) \cdot \mathbf{s}(a_n)}{\|\mathbf{s}(t)\|_2 \|\mathbf{s}(a_n)\|_2}, 
      \end{equation}
      attributes $a_n$ with $\text{SIM}\left(\mathbf{s}(t), \mathbf{s}(a_n)\right)$ smaller than $T_{sim}$ are deemed rarely relevant and subsequently discarded. With the aforementioned two steps of filtering irrelevant attributes, the remaining attributes are used to finetune a ResNet-152 classifier. 
    \end{enumerate}
  
  \setlength{\tabcolsep}{1pt}
  \begin{table}
  \centering
        \caption{Attribute Pipeline: accuracies based purely on $p_2$, comparing $3$ representation and classification strategies.}
        \label{tab:encoding_classification}
        \begin{tabular}{|l|c|c|}
          \hline
          Strategy & UCF101(s1) &  HMDB51(s1) \\
          \hline
          ResNet-152+Mean-Pool+FC & 20.09\% & 17.22\% \\
          ResNet-152+NetVLAD+FC & 20.10\% & 17.05\% \\
          Word2Vec+ResNet-152 finetune & \textbf{37.10\%} & \textbf{27.05\%} \\
          \hline
        \end{tabular}
  \end{table}
  
  The results of these three strategies are shown in Table~\ref{tab:encoding_classification}. For strategy \textit{ResNet-152 + Mean-Pool + FC} and \textit{ResNet-152 + NetVLAD + FC}, a pretrained (on ImageNet) ResNet-152 model is incorporated to extract features from cropped visual attribute candidates at the last ResNet ``avgpool" layer. For strategy \textit{ Word2Vec + ResNet-152 finetune}, the classifier ResNet-152 network is finetuned on cropped visual attribute candidates as follows. Regular SGD solver is used for optimization, with the initial learning rate of $0.001$, the learning rate decay of $0.1\times$ every 10 epochs, the momentum at $0.7$, the weight decay of $0.0005$ and a maximum of 20 epochs for training. Performance comparison of these $3$ strategies are summarized in Table~\ref{tab:encoding_classification}, where the strategy \textit{Word2Vec + ResNet-152 finetune} outperforms the other two by a substantial margin on both datasets. We inspected feature vectors extracted from the ResNet-152 component of both \textit{ResNet-152+Mean-Pool+FC} and \textit{ResNet-152+NetVLAD+FC} and discovered that these features are too similar, which could leads to the subpar performances. Therefore, the strategy \textit{Word2Vec+ResNet-152 finetune} are used for all subsequent experiments.  
  \end{appendices}
  \end{document}